LETTER

# Fast Computation of Moore-Penrose Inverse Matrices

Pierre Courrieu

Laboratoire de Psychologie Cognitive, UMR CNRS 6146, Université de Provence - Centre St Charles
bat. 9, case D, 3 place Victor Hugo, 13331 Marseille cedex 1, France.
E-mail: courrieu@up.univ-mrs.fr

*Abstract*- Many neural learning algorithms require to solve large least square systems in order to obtain synaptic weights. Moore-Penrose inverse matrices allow for solving such systems, even with rank deficiency, and they provide minimum-norm vectors of synaptic weights, which contribute to the regularization of the input-output mapping. It is thus of interest to develop fast and accurate algorithms for computing Moore-Penrose inverse matrices. In this paper, an algorithm based on a full rank Cholesky factorization is proposed. The resulting pseudoinverse matrices are similar to those provided by other algorithms. However the computation time is substantially shorter, particularly for large systems.

*Keywords*- Rank Deficient Least Square Systems, Moore-Penrose Inverse, Pseudoinverse, Generalized Inverse, Neural Learning, Minimum-norm Synaptic Weight Vectors, Regularization.

## 1. Introduction

The minimization of quadratic error functions is commonly used in learning algorithms of feed-forward neural networks. This can be achieved by applying gradient methods, but one can also use other methods of linear algebra and matrix computation, particularly in the case of Radial Basis Function Networks [1, 2]. Given a set of $m$ learning examples and a set of $n$ basis functions (or hidden neurons), with $m \geq n$, one forms the $m \times n$ real matrix **G** of the $n$ basis functions for the input values of the $m$ examples, and one must find a synaptic weight matrix **W** such that the quantity $\|\mathbf{GW} - \mathbf{F}\|^2$ is minimum, where **F** is the $m \times k$ real matrix of the expected output values (in $R^k$) for the $m$ examples. Whenever **G** is rank deficient, the solution (**W**) to the above minimization problem is not unique, it depends on the method used to solve the least square system, and possibly on some random variables (initial conditions, search parameters) for certain algorithms. However, one knows that the network input-output mapping must be suitably regularized in order to obtain a good generalization power, and the regularization is usually achieved by minimizing a norm of some differential operator applied to the mapping [1, 2]. In order to do this, one must choose suitable basis functions. However the choice of the synaptic weights of the output neurons can also play an important role. Let $K$ be a compact subset of $R^d$, on which a network provides an approximation $f$ of a sampled function. Then, for any input point $\mathbf{X} \in K$, the output $f(\mathbf{X})$ is given by the scalar product of the vector $g(\mathbf{X})$ of the $n$ basis function at **X** with the vector of output synaptic weights **w**. Let $\partial$ denote, for convenience, a partial derivative operator (of any order). One has

$$\sup_{\mathbf{X} \in K} |\partial f(\mathbf{X})| = \sup_{\mathbf{X} \in K} |\partial(g(\mathbf{X}).\mathbf{w})| \leq \|\mathbf{w}\| \sup_{\mathbf{X} \in K} \|\partial g(\mathbf{X})\| ,$$

which shows the interest of minimizing the norm $\|\mathbf{w}\|$ of the synaptic weight vector.

The Moore-Penrose inverse [3], also called Pseudoinverse, or Generalized Inverse, allows for solving least square systems, even with rank deficient matrices, in such a way that each column vector of the solution has a minimum norm, which is the desired property stated above. The Moore-Penrose inverse of a $m \times n$ matrix **G** is the unique $n \times m$ matrix $\mathbf{G}^+$ with the following four properties:

$$\mathbf{G}\mathbf{G}^+\mathbf{G} = \mathbf{G}, \quad \mathbf{G}^+\mathbf{G}\mathbf{G}^+ = \mathbf{G}^+, \quad (\mathbf{G}\mathbf{G}^+)' = \mathbf{G}\mathbf{G}^+, \quad (\mathbf{G}^+\mathbf{G})' = \mathbf{G}^+\mathbf{G},$$

where $\mathbf{M}'$ denotes the transpose (real case) or the adjoint (complex case) of the matrix $\mathbf{M}$. The desired least square solution is given by

$$\mathbf{W} = \mathbf{G}^+ \mathbf{W}.$$

Whenever $\mathbf{G}$ is of full rank ($n$), the Moore-Penrose inverse reduces to the usual pseudoinverse:

$$\mathbf{G}^+ = (\mathbf{G}'\mathbf{G})^{-1} \mathbf{G}'.$$

However, when $\mathbf{G}$ is rank deficient (that is when its rank $r$ is lower than $n$), the computation of $\mathbf{G}^+$ is more complex. There are several methods for computing Moore-Penrose inverse matrices [3]. The most commonly used is the Singular Value Decomposition (SVD) method, that is implemented, for example, in the "*pinv*" function of Matlab (version 6.5.1), as well as in the "*PseudoInverse*" function of Mathematica (version 5.1). This method is very accurate, but time consuming when the matrix is large. Other well-known methods are Greville's algorithm, the full rank QR factorization by Gram-Schmidt orthonormalization (GSO), and iterative methods of various orders [3]. A number of expansions of the Moore-Penrose inverse can also be used to develop effective algorithms [4], while certain algorithms are specially devoted to parallel computation [5], but these last ones are hardly usable on serial processors given that they use Cramer's rule.

In this paper, we propose an algorithm for fast computation of Moore-Penrose inverse matrices on any computer. The algorithm is based on a known reverse order law (eq. 3.2 from [4]), and on a full rank Cholesky factorization of possibly singular symmetric positive matrices (Theorem 4 from [6]).

## 2. Algorithm Foundations

Consider the symmetric positive $n \times n$ matrix $\mathbf{G}'\mathbf{G}$, and assume that it is of rank $r \leq n$. Using Theorem 4 from [6], one knows that there is a unique upper triangular matrix $\mathbf{S}$ with exactly $n-r$ zero rows, and such that $\mathbf{S}'\mathbf{S} = \mathbf{G}'\mathbf{G}$, while the computation of $\mathbf{S}$ is a simple extension of the usual Cholesky factorization of non-singular matrices. Removing the zero rows from $\mathbf{S}$, one obtains a $r \times n$ matrix, say $\mathbf{L}'$, of full rank $r$, and we have

$$\mathbf{G}'\mathbf{G} = \mathbf{S}'\mathbf{S} = \mathbf{L}\mathbf{L}' \tag{1}$$

Note that one can as well directly compute the matrix $\mathbf{L}$, as this is implemented, in Matlab code, in the function "*geninv*" listed in Appendix.

We use also a general relation concerning the Moore-Penrose inverse of a matrix product $\mathbf{AB}$ given by eq. 3.2 from [4]

$$(\mathbf{AB})^+ = \mathbf{B}'(\mathbf{A}'\mathbf{A}\mathbf{B}\mathbf{B}')^+ \mathbf{A}' \tag{2}$$

With $\mathbf{B} = \mathbf{I}$, one obtains from (2)

$$\mathbf{A}^+ = (\mathbf{A}'\mathbf{A})^+ \mathbf{A}' \tag{3}$$

If $\mathbf{B} = \mathbf{A}'$ and $\mathbf{A}$ is a $n \times r$ matrix of rank $r$, then one obtains from (2)

$$(\mathbf{A}\mathbf{A}')^+ = \mathbf{A}(\mathbf{A}'\mathbf{A})^{-1}(\mathbf{A}'\mathbf{A})^{-1}\mathbf{A}'. \tag{4}$$

We are now ready to prove the following result:

**Theorem 1.** With $\mathbf{G}$ and $\mathbf{L}$ defined as above, one has $\mathbf{G}^+ = \mathbf{L}(\mathbf{L}'\mathbf{L})^{-1}(\mathbf{L}'\mathbf{L})^{-1}\mathbf{L}'\mathbf{G}'$.

*Proof.* Using Eq.(3) one obtains

$$\mathbf{G}^+ = (\mathbf{G}'\mathbf{G})^+ \mathbf{G}',$$

and using Eqs. (1) and (4), one obtains

$$(\mathbf{G}'\mathbf{G})^+ = (\mathbf{L}\mathbf{L}')^+ = \mathbf{L}(\mathbf{L}'\mathbf{L})^{-1}(\mathbf{L}'\mathbf{L})^{-1}\mathbf{L}',$$

which completes the proof. □

The function "*geninv*" listed in Appendix provides all implementation details of the above solution, in Matlab code, and it returns the Moore-Penrose inverse of any rectangular matrix given as argument. The two main operations are the full rank Cholesky factorization of $\mathbf{G'G}$ and the inversion of $\mathbf{L'L}$. On a serial processor, these operations are of complexity order $O(n^3)$ and $O(r^3)$, respectively. However, in a parallel architecture, with as many processors as necessary, the time complexity for the Cholesky factorization of $\mathbf{G'G}$ could reduce to $O(n)$, as suggested by the notation in Matlab code, while the time complexity for the inversion of the symmetric positive definite matrix $\mathbf{L'L}$ could reduce to $O(\log r)$, according to [7].

## 3. Computational Test

In this section, we compare the performance of the proposed method ("*geninv*" function) to that of four usual algorithms for the computation of Moore-Penrose inverse matrices: Greville's method [8], the SVD method (Matlab *pinv* function), the full rank QR factorization by GSO method, and an iterative method of optimized order [3]. For this last one, the optimal order was determined experimentally, and it was found to be $p = 2^9 = 512$ for the family of matrices that were tested. All algorithms were carefully implemented and tested in Matlab (Version 6.5.1) on a common personal computer. The test matrices were of size $m \times n$, with $n = 2^k$, $k = 5..10$, and $m = 2n$, and they were rank deficient, with rank $r = 7n/8$. The matrix coefficients were real random numbers in [-1,1]. The accuracy of the algorithms was examined in error matrices corresponding to the four properties characterizing the Moore-Penrose inverse: $\mathbf{GG^+G - G}$, $\mathbf{G^+GG^+ - G^+}$, $\mathbf{(GG^+)' - GG^+}$, and $\mathbf{(G^+G)' - G^+G}$. It turned out that, for all algorithms and for all test matrices, none of the coefficients in the error matrices was greater than $2 \times 10^{-10}$ in absolute value, thus we obtained a reliable approximation of the Moore-Penrose inverse in all cases. The computation times (in seconds) are reported in Table 1, where one can see that the *geninv* function is substantially faster than the other algorithms for large matrices. An exception is observed for the smallest matrices ($n = 32$), for which the full rank QR method is a bit faster than the *geninv*. This probably results from the use of square root functions in the full rank Cholesky factorization of the *geninv*, while GSO uses only basic arithmetic operations. However, this small disadvantage is rapidly compensated by other properties as the size of the test matrices increases.

**Table 1.** Computation time (in seconds) of the Moore-Penrose inverse of random rectangular rank-deficient real matrices by four usual algorithms and the "*geninv*" function, as a function of the matrix size. The computation error is lower than $2 \times 10^{-10}$ per coefficient in the error matrices, in all cases.

|            | Greville's method | SVD method (Matlab pinv) | full rank QR (by GSO) | Iterative (of order 512) | geninv  |
|------------|-------------------|--------------------------|-----------------------|--------------------------|---------|
| *n = 32*   | 0.015             | 0.025                    | 0.006                 | 0.019                    | 0.007   |
| *n = 64*   | 0.296             | 0.044                    | 0.035                 | 0.141                    | 0.022   |
| *n = 128*  | 1.734             | 0.363                    | 0.453                 | 0.664                    | 0.105   |
| *n = 256*  | 17.092            | 5.433                    | 3.795                 | 4.158                    | 0.768   |
| *n = 512*  | 169.850           | 49.801                   | 29.197                | 29.286                   | 6.445   |
| *n = 1024* | 2160.100          | 373.895                  | 240.220               | 232.688                  | 54.137  |

## 4. Application Fields

As stated above, the proposed tool allows for fast computation of Moore-Penrose inverse matrices and fast solving of large least square systems, possibly with rank deficient matrices. In the case of rank deficiency, the obtained solution, in a learning process, has the advantage of being of minimum norm, and thus it contributes to the regularization of the input-output mapping. However, we must note that whenever there is no risk of rank deficiency, there are simpler ways of solving least square systems, and the Moore-Penrose inverse itself reduces to a simpler expression. Thus, the choice of using the proposed tool must depend on the possibility that the matrix $\mathbf{G}$ be rank deficient, which in fact depends on the choice of the set of basis functions [9]. One can say that the Moore-Penrose inverse extends modeler's freedom concerning this last point. Moreover, we can note that, without being actually rank deficient, the matrix $\mathbf{G}$ can have certain column vectors that are close to be linearly dependent, in such a way that the least square system is ill-conditioned. In this case, the use of the Moore-Penrose inverse avoids catastrophic numerical results that can result from the ill-conditioning. Another characteristic of the proposed tool is that it is particularly fast, which is critical whenever there are time constraints, as in on-line learning, or whenever the computation must be updated many times, as in incremental learning procedures. In this last case, there are known fast updating procedures [10], however these procedures do not support rank deficiency, and the basis functions cannot be modified after they have been included in the network, while, in fact, it is frequently desirable to rescale Radial Basis Functions, for example, as the size of the network increases. Thus, it is clear that the fast computation of Moore-Penrose inverse matrices can have valuable applications in neurocomputational learning procedures.

### Appendix: Matlab code of the "geninv" function

```
function Y = geninv(G)
% Returns the Moore-Penrose inverse of the argument
    % Transpose if m < n
[m,n]=size(G); transpose=false;
if m<n
    transpose=true;
    A=G*G';
    n=m;
else
    A=G'*G;
end
    % Full rank Cholesky factorization of A
dA=diag(A); tol= min(dA(dA>0))*1e-9;
L=zeros(size(A));
r=0;
for k=1:n
    r=r+1;
    L(k:n,r)=A(k:n,k)-L(k:n,1:(r-1))*L(k,1:(r-1))';
    % Note: for r=1, the substracted vector is zero
    if L(k,r)>tol
        L(k,r)=sqrt(L(k,r));
        if k<n
            L((k+1):n,r)=L((k+1):n,r)/L(k,r);
        end
    else
        r=r-1;
    end
end
L=L(:,1:r);
    % Computation of the generalized inverse of G
M=inv(L'*L);
if transpose
    Y=G'*L*M*M*L';
else
    Y=L*M*M*L'*G';
end
```

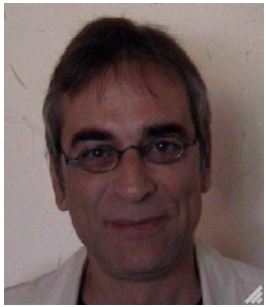

**Pierre Courrieu** is a CNRS researcher currently working at the University of Provence (France) with psychologists and neuroscientists. He is a member of the European Neural Network Society, and has published works on visual shape recognition, neural computation, data encoding, multidimensional scaling, function approximation on metric and non-metric spaces, and global optimization methods.